\documentclass[10pt,twocolumn,letterpaper]{article}

\usepackage{cvpr}              

\usepackage[dvipsnames]{xcolor}

\usepackage{setspace}

\usepackage[accsupp]{axessibility}
\definecolor{cvprblue}{rgb}{0.21,0.49,0.74}
\usepackage[pagebackref,breaklinks,colorlinks,citecolor=cvprblue]{hyperref}

\def\paperID{1853} 
\def\confName{CVPR}
\def\confYear{2024}

\title{Learning Background Prompts to Discover Implicit Knowledge for Open Vocabulary Object Detection}

\author{Jiaming Li$^{1}$ \quad
Jiacheng Zhang$^{1}$\quad
Jichang Li$^{1,2}$\quad
Ge Li$^{3}$\quad 
Si Liu$^{4}$\quad 
\\
Liang Lin$^{1}$\quad
Guanbin Li $^{1,5,6}$\thanks{Corresponding author.}\\
$^{1}$School of Computer Science and Engineering, Sun Yat-sen University, Guangzhou, China\\
$^{2}$Department of Computer Science, The University of Hong Kong, Hong Kong\\
$^{3}$SECE, Shenzhen Graduate School, Peking University, Shenzhen, China\\
$^{4}$Institute of Artificial Intelligence,
Beihang University, China\\
$^{5}$GuangDong Province Key Laboratory of Information Security Technology\\
$^{6}$Research Institute, Sun Yat-sen University, Shenzhen, China\\
{\tt\small \{lijm48,zhangjch58\}@mail2.sysu.edu.cn, \{liguanbin,linlng\}@mail.sysu.edu.cn }\\
{\tt\small csjcli@connect.hku.hk, lige@pku.edu.cn, liusi@buaa.edu.cn}
}

\newcommand{\nothing}[1]{}

\definecolor{DeltaColor}{rgb}{0.039,0.73,0.71}
\definecolor{SetaColor}{rgb}{0.867, 0.0235, 0.376}
\definecolor{SigmaColor}{rgb}{0.98,0.45,0.0}
\definecolor{RedColor}{rgb}{0.8,0,0}
\definecolor{AlphaColor}{rgb}{0,0,0.8}
\definecolor{BetaColor}{rgb}{0.8,0,0.8}
\definecolor{GammaColor}{rgb}{0.5,0,0.7}
\definecolor{EpsilonColor}{rgb}{0.353,0.725,0.906}
\definecolor{TauColor}{rgb}{0.423,0.235,0.192}
\definecolor{WtColor}{rgb}{0.235,0.470,0.470}

\definecolor{AudioColor}{rgb}{0.56,0.34,0.62}

\definecolor{DeadlineColor}{rgb}{0.9,0.4,0} 

\definecolor{figred}{rgb}{1,0,0}
\definecolor{figgreen}{rgb}{0,0.6,0}
\definecolor{figblue}{rgb}{0,0,1}
\definecolor{figpink}{rgb}{1,0.63,0.63}

\newcounter{pccount}
\newcommand{\filename}[1]{\url{#1}}
\newcommand{\foldername}[1]{\url{#1}}

\begin{document}
\maketitle

\begin{abstract}
Open vocabulary object detection (OVD) aims at seeking an optimal object detector capable of recognizing objects from both base and novel categories. Recent advances leverage knowledge distillation to transfer insightful knowledge from pre-trained large-scale vision-language models to the task of object detection, significantly generalizing the powerful capabilities of the detector to identify more unknown object categories. However, these methods face significant challenges in background interpretation and model overfitting and thus often result in the loss of crucial background knowledge, giving rise to sub-optimal inference performance of the detector. To mitigate these issues, we present a novel OVD framework termed LBP to propose learning background prompts to harness explored implicit background knowledge, thus enhancing the detection performance w.r.t. base and novel categories. Specifically, we devise three modules: Background Category-specific Prompt, Background Object Discovery, and Inference Probability Rectification, to empower the detector to discover, represent, and leverage implicit object knowledge explored from background proposals. Evaluation on two benchmark datasets, OV-COCO and  OV-LVIS, demonstrates the superiority of our proposed method over existing state-of-the-art approaches in handling the OVD tasks.
\end{abstract}

\section{Introduction}

Compared to conventional vision tasks~\cite{li2023betweenness, li2023idm, li2024feddiv, li2021cross, li2022nce, zhang2023semidetr, huang2023divide, AlignSAM}, object detection has witnessed significant success in research, such as~\cite{ren2015faster, lin2017feature, shao2019objects365, zheng2022towards}, however, solely detecting and classifying objects within known categories (base classes) during inference significantly diminishes its generalization capacity in real-world applications. Open Vocabulary Object Detection (OVD) emerges as a prospective means to overcome this restriction, endowed with the capability to detect unseen categories (novel classes), without the explicit need for annotations.

Leveraging large-scale pre-trained Vision and Language Models (PVLMs), exemplified by CLIP~\cite{radford2021learning} and ALIGN~\cite{jia2021scaling}, recent advances in OVD, e.g.~\cite{gu2021open, du2022learning, wu2023aligning}, employ knowledge distillation to transfer insightful knowledge of PVLMs to the task of object detection, generalizing its powerful capabilities to detect more unknown object categories. 
However, these methods face a critical limitation in background interpretation. Specifically, these approaches tend to represent background proposals with a single ``background class'', encompassing all categories beyond foreground base classes. With solely a singular learnable embedding to interpret this background class, the trained model may fail to capture the diverse implicit knowledge within background proposals. 
This leads to the loss of essential background information, resulting in incomplete and ambiguous representations of unseen categories by the model. Consequently, the model is incapable of distinguishing objects from those unknown categories using their representations.

\begin{figure}[t]
\centering
\includegraphics[width=0.95\linewidth, trim=0 10 0 0, clip]{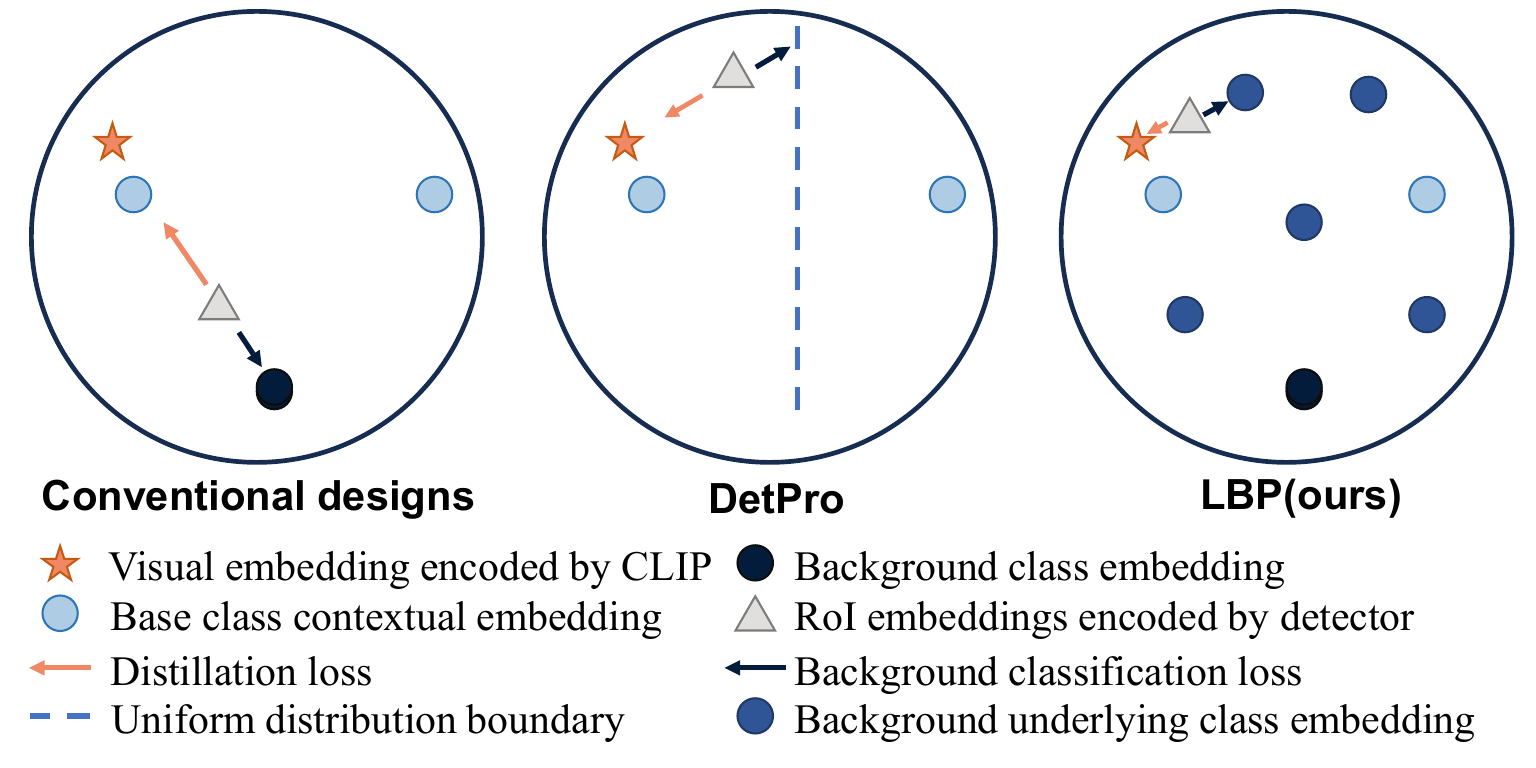}
\caption{An example to illustrate previous and our designs in background interpretation. 
Conventional designs use a single background embedding to push the RoI embedding away from the CLIP embedding. DetPro~\cite{du2022learning} proposes to uniformly push the RoI embedding away when the CLIP embedding nears a base class embedding, leading to a loss of class relation. Our LBP, on the other hand, learns multiple background underlying class embeddings, effectively preserving class relations and alleviating loss conflict.
``Distillation loss'' uses knowledge distillation to align visual features encoded by the decoder with CLIP embeddings, while ``Background classification loss'' refers to the classification loss for background proposals.
}
\vspace{-0.20cm}
\label{fig:teaser}
\end{figure}

To tackle this problem, researchers experimented with various strategies like binary cross-entropy loss~\cite{bangalath2022bridging} or soft background loss~\cite{du2022learning} for background interpretation.
However, their basic assumption that background proposals are uniformly dissimilar to any foreground category overlooks the nuanced class relations between foreground and background classes, as illustrated in Figure~\ref{fig:teaser}. Additionally, these methods struggle with model overfitting during training due to an abundance of fully supervised data from base classes, which makes detectors biased toward those categories.
  Recent advances like \cite{zhao2022exploiting,bangalath2022bridging,gao2022open,feng2022promptdet} propose mining implicit objects of novel categories from background proposals and employing pseudo-labeling to enhance their interpretation. Yet, these methods often rely on additional prior knowledge, such as the names of novel categories,  leaving substantial unexplored knowledge.

In this paper, we present LBP, a novel framework for open-vocabulary object detection.
Without any prior knowledge, LBP proposes learning background prompts to harness explored implicit background knowledge, thus enhancing the detection performance w.r.t. base and novel categories. Specifically, the LBP framework initially introduces a Background Category-specific Prompt module.  It discovers and represents background underlying categories estimated from background proposals by leveraging learnable category-specific contexts, consequently resulting in improved background interpretation. Then, an online module, namely Background Object Discovery, is introduced to further explore and exploit implicit object knowledge correlated with those estimated underlying categories from background proposals, significantly contributing to mitigating model overfitting. 
Moreover, an Inference Probability Rectification module is presented to address conceptual overlaps between estimated background categories and novel categories provided during inference. This rectification enables the model to accurately compute probabilities for novel categories, thereby significantly enhancing the detector performance. The contributions are summarized as follows.
\begin{itemize}
    \item We propose a novel framework, termed LBP, for open-vocabulary object detection, where learning background prompts is presented to harness explored implicit background knowledge, thereby enhancing detection of both base and novel categories during inference.
    
    \item 
    We devise three modules: Background Category-specific Prompt, Background Object Discovery, and Inference Probability Rectification,  to empower the detector to discover, represent, and leverage implicit object knowledge explored from background proposals.
    
    \item 
    Evaluation on two benchmark datasets, OV-COCO \cite{lin2014microsoft} and  OV-LVIS \cite{gupta2019lvis}, demonstrates the superiority of our proposed method over existing state-of-the-art approaches in handling the OVD tasks.
\end{itemize}

\begin{figure*}[t]
\centering
\includegraphics[width=\textwidth, height=9.0 cm, trim=10 0 20 5, clip]{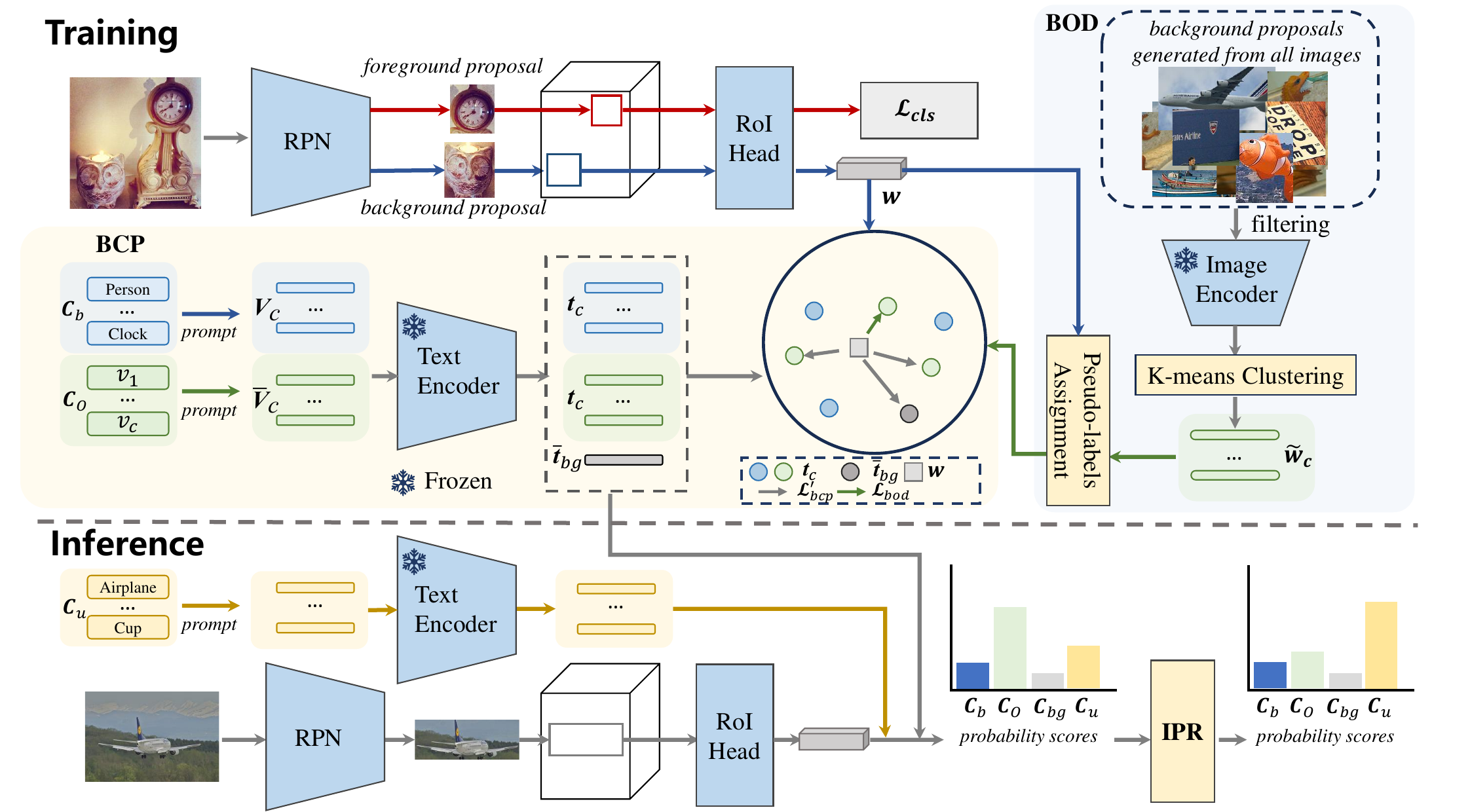}
\caption{An overview of the proposed LBP approach, consisting of three modules:  Background Category-specific Prompt (BCP), Background Object Discovery (BOD), and Inference Probability Rectification (IPR).
During training, BCP is first proposed to discover and represent background underlying categories, estimated from background proposals, with learnable background category-specific contexts. Then, BOD is presented to employ $k$-means clustering on background proposals across all images to harness implicit objects explored from background underlying knowledge. During inference, IPR is introduced to rectify probability scores of novel categories provided, by loosening their conceptual overlaps with background underlying categories estimated from background proposals.
}
\label{fig:pipeline}
\end{figure*}

\section{Related Work}

\textbf{Open vocabulary object detection (OVD).} 
OVR-CNN~\cite{zareian2021open} pioneered open vocabulary object detection, by expanding object detection capabilities through image-caption datasets. 
Subsequently, prevailing approaches to tackle OVD have primarily relied on fine-tuning PVLMs~\cite{kim2023region,minderer2022simple,zhong2022regionclip,li2022grounded,kuo2022f,lin2022learning}. However, retraining PVLMs with large-scale learnable model weights is resource-intensive. 
Consequently, more recent approaches rooted in knowledge distillation~\cite{wang2023object,ma2022open,wu2023cora} to address OVD by leveraging knowledge distilled from CLIP~\cite{radford2021learning} into the detector, predominantly adopting source-free strategies. For example, ViLD~\cite{gu2021open} aligns detector-encoded features with CLIP embeddings, while BARON~\cite{wu2023aligning} generates region ensembles using random masking within grid spaces to distill visual concept co-occurrence within a scene.
Despite partially transferring knowledge of novel categories from PVLMs, detectors exhibit bias towards base classes due to disparities between vision-language alignment and detection tasks. To mitigate this bias, researchers explore pseudo labeling techniques utilizing prior knowledge of novel category names~\cite{zhao2022exploiting} or weakly-supervised annotations to the detection dataset~\cite{zhou2022detecting,bangalath2022bridging,gao2022open}. However, integrating additional knowledge poses limitations in real-world scenarios.

Our proposed method extracts background implicit knowledge about classes beyond known/base categories from the provided detection dataset without requiring prior information about novel categories during inference. This utilization of implicit knowledge enables the detector to identify objects from unknown categories, enhancing its applicability across diverse scenarios.



\textbf{Prompt tuning for OVD:}
Prompt tuning, stemming from natural language processing, has evolved with PVLM advancements to enhance their efficacy in specific downstream tasks~\cite{khattak2023maple}. CoOp, proposed in \cite{zhou2022learning}, introduces learnable contexts for 2D vision-language classification. CoCoOp, an extension proposed by~\cite{zhou2022conditional}, further enhances CoOp by generating input-conditioned tokens for each image. Various visual prompting approaches advocate integrating prompts into the image encoder of PVLMs, including works such as \cite{jia2022visual,bahng2022visual,guo2023viewrefer,chen2023visual}.

In the realm of OVD, DetPro~\cite{radford2021learning} pioneered prompt engineering by implementing a soft background loss, significantly improving the detector's ability to represent specific categories. Similarly, proposals for pre-training prompts on large-scale datasets~\cite{ren2023prompt,feng2022promptdet} aim to generate more universally applicable prompts.
In contrast to DetPro~\cite{radford2021learning}, our approach has no reliance on uniform distribution assumption. Instead, by prompting underlying background categories estimated from proposals, our method enhances the learning of unknown category representations, leading to significant performance gains in detection.

\section{Approach}



\subsection{Preliminaries}

\noindent
\textbf{Problem formulation.}
Open vocabulary object detection (OVD) aims at seeking an optimal object detector capable of recognizing objects from both base and novel (previously unseen) categories, denoted by $\mathcal{C}_b=\{c_i\}_{i=1}^{n_b}$ and $\mathcal{C}_u=\{c_i\}_{i=1}^{n_u}$, respectively, in the inference dataset $\mathcal{D}_I$. This detector is achieved through optimization on the training dataset $\mathcal{D}_T$, annotated exclusively with instances from the base classes within $\mathcal{C}_b$.
Here, $\mathcal{C}_b \cap \mathcal{C}_u = \emptyset$, while $n_b$ and $n_u$ represent the cardinality of the respective category sets.


Built upon Faster R-CNN~\cite{ren2015faster}, given an input image, it is initially encoded by Faster R-CNN into global image features and generates a set of proposals through a region proposal network (RPN).
In general, these region proposals are categorized into \textit{foreground} proposals and \textit{background} proposals~\cite{du2022learning}, designated as $\mathcal{P}$ and $\mathcal{N}$, respectively. The {foreground} proposals in $\mathcal{P}$ encompass all base categories within $\mathcal{C}_b$, while all {background} proposals from $\mathcal{N}$ are collectively classified as a singular \textbf{\textit{super-class}} (denoted as $\textit{\textbf{c}}_{bg}$ here), representing any category outside of $\mathcal{C}_b$.



\noindent
\textbf{Knowledge distillation.}
Leveraging large-scale pre-trained large-scale Vision and Language Models (PVLMs), specifically CLIP~\cite{zhou2022learning}, knowledge distillation empowers Faster R-CNN to effectively detect objects from arbitrary vocabularies by transferring insightful knowledge from CLIP. This process involves aligning the proposal features generated by the detector (actually output by the RoI Head of the detector, denoted by  $\textit{\textbf{w}}{(\cdot)}$) with those extracted by the image encoder of CLIP (denoted by  $\mathcal{I}{(\cdot)}$), effectively simulating coherence between the feature spaces of the detector and CLIP. Consequently, this process implicitly projects the visual embedding space of the detector into the textual embedding space of CLIP (output by the text encoder of CLIP, denoted by  $\mathcal{T}{(\cdot)}$). In this way,  the original classifier of Faster R-CNN is replaced.



\noindent
\textbf{Foreground/background interpretation.}
Given class $c$ from $\mathcal{C}_b$, its contextual prompt is defined as \texttt{`a photo of \{category\} in the scene'}, noted as $\textit{\textbf{V}}_c$. Here,  \texttt{`\{category\}'} denotes the contextual word for class $c$. Then its contextual embedding is obtained by feeding $\textit{\textbf{V}}_c$ into the CLIP text encoder $\mathcal{T}(\cdot)$, as follows,
\begin{equation}
\label{eq:context_embed_foreground}
\textit{\textbf{t}}_{c} = \mathcal{T}(\textit{\textbf{V}}_c).
\end{equation}
However, it is challenging to determine the contextual word for the ``\textit{background}'' class $\textit{\textbf{c}}_{bg}$. ViLD~\cite{gu2021open} or BARON~\cite{wu2023aligning} represents it with a learnable contextual embedding, without encoding it via CLIP's text encoder $\mathcal{T}(\cdot)$, thus defining it as $\textit{\textbf{t}}_{bg}$.


\noindent
\textbf{Probability calculation.}Having obtained $\{\textit{\textbf{t}}_{c}|c\in\mathcal{C}_{b}\}\cup\{\textit{\textbf{t}}_{bg}\}$, let $\textit{\textbf{w}}{(\cdot)}$  denote the visual embedding of a proposal generated by the RoI Head of the object detector. Consider $\mathcal{C}_b^{bg}=\mathcal{C}_b\cup\{\textit{\textbf{c}}_{bg}\}$. For any proposal $x\in\mathcal{P}\cup\mathcal{N}$ from $\mathcal{D}_T$ with its visual embedding $\textit{\textbf{w}}{(x)}$, the probability of categorizing this proposal under the category $c\in\mathcal{C}_b^{bg}$ is defined as follows:
\begin{equation}
\label{prob}
p(c|x; \mathcal{C}_b^{bg}) = \frac{\exp( \cos(\textit{\textbf{w}}(x), \textit{\textbf{t}}_{c})/ \tau)}{\sum_{c^{\prime}\in \mathcal{C}_b^{bg}} \exp(\cos(\textit{\textbf{w}}(x), \textit{\textbf{t}}_{c^{\prime}})/ \tau)},
\end{equation}
where $\exp(\cdot)$ and $\cos(\cdot)$ represent the exponential function and cosine similarity, respectively, while $\tau$ is the temperature parameter used for rescaling the values. 

During inference, once the novel classes from $\mathcal{C}_u$ are introduced, the probability of a proposal from $\mathcal{D}_I$ being classified as class $c$ of $\mathcal{C}_b\cup\{\textit{\textbf{c}}_{bg}\}\cup\mathcal{C}_u$ during detection can be computed simply by replacing $\mathcal{C}_b^{bg}$ in Eq. (\ref{prob}) with $\mathcal{C}_{b,u}^{bg}=\mathcal{C}_b\cup\{\textit{\textbf{c}}_{bg}\}\cup\mathcal{C}_u$.
\subsection{The Proposed Method}
In this paper, we introduce LBP, a novel framework designed for open-vocabulary object detection. The key idea of LBP is that without any prior knowledge, learning background prompts is proposed to harness explored implicit background knowledge. This ensures the model attains improved background interpretation and decreased model overfitting, thus empowering the detector capable of better recognizing both base and novel categories. Specifically, the LBP framework initially introduces a Background Category-specific Prompt (\textbf{BCP}) module, discovering underlying categories from background proposals, and representing those using learnable context-specific prompts. Moreover, we introduce an online Background Object Discovery (\textbf{BOD}) to further exploit implicit object knowledge w.r.t. those estimated underlying categories, consequently alleviating model bias towards base categories, and present an Inference Probability Rectification (\textbf{IPR}) module to resolve conceptual overlaps between estimated background categories and novel categories during inference, leading to precisely computed probability scores for novel categories.
Our primary focus lies in exploring background interpretation, distinct from prevailing OVD approaches rooted in knowledge distillation. This enables seamless integration of our method into existing OVD frameworks~\cite{wu2023aligning} and ViLD~\cite{gu2021open}, etc.
An overview of the proposed approach is illustrated in Figure~\ref{fig:pipeline}.

\subsubsection{Background Category-specific Prompt}
\label{sec:BCP}
In this section, we model the underlying categories for background proposals in OVD and learn the corresponding category-specific prompts. 
However, the absence of prior knowledge of those categories necessitates the estimation of their optimal number in background proposals. 

Towards this end, we first produce background proposals for all training images using a class-agnostic RPN trained on $\mathcal{D}_T$ with only base classes, adopting a similar technique as in VL-PLM~\cite{zhao2022exploiting}. 
Those proposals are subsequently filtered based on RPN scores (with a threshold ${\theta}$), and Non-Maximum Suppression (NMS). 
Then, we obtain features for these filtered proposals generated from the image encoder of CLIP,
and we refer to~\cite{han2019learning,zheng2022towards} to estimate the optimal number of the underlying categories in background proposals via K-mean clustering~\cite{vaze2022generalized}, denoted by $n_o$, in the obtained proposal features.
During this procedure, potential ambiguity in data partitioning during estimation may arise due to feature overlap, leading to the underestimation of $n_o$. Consequently, we expand the count of background underlying categories to $n_o+n_a$, where $n_a$ serves as a pre-defined hyper-parameter. Here, let $\mathcal {C}_O=\{c_i\}_{i=1}^{n_o+n_a}$ denote those estimated background categories, for simplicity.

However, due to the absence of prior knowledge of $\mathcal {C}_O$, identifying their contextual words is a non-trivial task. To address this issue, we propose to learn category-specific context vectors.
In this case, the contextual words of those background underlying categories can be described with the learned continuous vectors. 
Specifically, we adopt the prompt form for a given class  $c\in \mathcal{C}_O$ as follows:
\begin{equation}
\overline{\textit{\textbf{V}}}_c = \text{\texttt{`a photo of }} \{{v}_c\} \text{\texttt{ in the scene'}},
\end{equation}
where the context vectors $\{{v}_c|{c\in\mathcal{C}_O}\}$ are learnable.
Following Eq. (\ref{eq:context_embed_foreground}), the corresponding contextual embeddings can be acquired as $\overline{\textit{\textbf{t}}}_{c} = \mathcal{T}(\overline{\textit{\textbf{V}}}_c)$.

Additionally, we present a specific \textit{\textbf{sub-class}} category, $\overline{\textit{\textbf{c}}}_{bg}$, designed to represent the ``background'' within background proposals, distinct from the super-class ``\textit{background}'' ${\textit{\textbf{c}}}_{bg}$. Specifically, $\overline{\textit{\textbf{c}}}_{bg}$ covers unexplored categories not estimated in $\mathcal{C}_O$ and includes any \textit{\textbf{untargeted}} object within background proposals. Similar to $\textit{\textbf{t}}_{bg}$, we define a learnable contextual embedding, $\overline{\textit{\textbf{t}}}_{bg}$, for this category $\overline{\textit{\textbf{c}}}_{bg}$.


Once we obtain $\{\textit{\textbf{t}}_{c}|c\in\mathcal{C}_{b}\}\cup\{\overline{\textit{\textbf{t}}}_{c}|c\in\mathcal{C}_{O}\}\cup\{\overline{\textit{\textbf{t}}}_{bg}\}$, the corresponding set of categories can be defined as $\mathcal{C}_{b,o}^{bg}=\mathcal{C}_b\cup\mathcal{C}_O\cup\{\overline{\textit{\textbf{c}}}_{bg}\}$. With reference to Eq. (\ref{prob}), 
for any proposal $x\in\mathcal{P}\cup\mathcal{N}$, its probability of being classified as class $c\in\mathcal{C}_{b,o}^{bg})$ can be defined as
$p(c|x; \mathcal{C}_{b,o}^{bg})$. At this point, the model optimization in the \textbf{BCP} for the proposals in $\mathcal{P}$ and $\mathcal{N}$ can be achieved using the following cross-entropy losses:
\begin{equation}
\label{eq:cross_entropy_foreground}
\mathcal{L}_{cls} =  \frac{1}{|\mathcal{P}|} \sum_{x\in\mathcal{P}}{ -\log p(c=\textit{\textbf{C}}(x)}|x; \mathcal{C}_{b,o}^{bg}),
\end{equation}
\begin{equation}
\label{eq:bh}
\mathcal{L}_{bcp} =  \frac{1}{|\mathcal{N}|} \sum_{x\in\mathcal{N}}{-\log \sum_{c\in{\mathcal{C}_O\cup\{\overline{\textit{\textbf{c}}}_{bg}}\}} p(c|x; \mathcal{C}_{b,o}^{bg})},
\end{equation}
 where $\textit{\textbf{C}}(x)$ represents the true label class of the given proposal $x$.
  $\mathcal{L}_{bcp}$ enhances the sum of probabilities of all background underlying categories estimated and $\overline{\textit{\textbf{c}}}_{bg}$ within each background proposal, enabling the detector to softly learn the probabilities scores assigned for those categories.

However, contextual embeddings of $\mathcal{C}_O\cup\{\overline{\textit{\textbf{c}}}_{bg}\}$ might not be sufficient to represent all background proposals with diverse scenes. When a background proposal belongs to $\overline{\textit{\textbf{c}}}_{bg}$ and cannot be adequately represented by its contextual embedding, the visual embeddings of that proposal will be distant from any contextual embeddings of the estimated categories from $\mathcal{C}_O$. Consequently, the sum of probabilities of those categories, i.e., $\sum_{c\in{\mathcal{C}_O\cup\{\overline{\textit{\textbf{c}}}_{bg}}\}} p(c|x; \mathcal{C}_{b,o}^{bg})$ in Eq. (\ref{eq:bh}), denoted as $\textit{\textbf{p}}_o^{bg}$, for simplicity,  would become very small.

In this situation, we should uniformly push the visual embeddings of that proposal towards the estimated background underlying categories and $\overline{\textit{\textbf{c}}}_{bg}$, given that the proposal is distinct from those categories.
When $\textit{\textbf{p}}_o^{bg} < \gamma$, we introduce an additional softer background loss component to relax $\mathcal{L}_{bcp}$ for all background proposals in $\mathcal{N}$ as follows,
\begin{equation}
\mathcal{L}_{rlx} =
\frac{1}{|\mathcal{N}|}\sum_{x\in\mathcal{N}} \frac{1}{n_{oa}} \sum_{c\in{\mathcal{C}_O\cup\{\overline{\textit{\textbf{c}}}_{bg}\}}}- \log p(c|x; \mathcal{C}_{b,o}^{bg}),
\end{equation}
where $n_{oa}=n_{o}+n_{a}+1$ represents the size of $\mathcal{C}_O\cup\{\overline{\textit{\textbf{c}}}_{bg}\}$, while $\gamma$ is a threshold with a small value.  Therefore, the final version of the loss component in this \textbf{BCP}  for background proposals to optimize the model is formulated as,
\begin{equation}
\label{eq:bcp}
\mathcal{L}_{bcp}^{\prime} =
\begin{cases}
\mathcal{L}_{bcp}, & \text{if}~\textit{\textbf{p}}_o^{bg} \geq \gamma, \\
\mathcal{L}_{rlx}, & \text{otherwise}.
\end{cases}
\end{equation}
Note that $\mathcal{L}_{bcp}^{\prime}$ enriches the discrimination capability of the detector by incorporating diverse contextual embeddings of estimated background categories, diverging from the soft background loss in DetPro~\cite{du2022learning}. This enables the model to exploit implicit object knowledge w.r.t. those estimated underlying categories, while also reconciling training conflicts between the losses utilized for classification and distillation, as highlighted in prior works~\cite{wu2023aligning}.
 


\subsubsection{Background Object Discovery}

To enhance model training, we introduce an online Background Object Discovery (\textbf{BOD}) module to effectively discover and exploit unseen objects. This aims to extract implicit objects w.r.t. the underlying categories estimated from background proposals.
To simplify, we here decompose $\mathcal{C}_O$ into $\mathcal{C}_O^{\prime} = \{c_i\}_{i=1}^{n_o}$ and $\mathcal{C}_a = \{c_i\}_{i=1}^{n_a}$ in this section.


During initial training, with a given $n_o$, we conduct $k$-means clustering (where $k$ is set to $n_o$) on the set of visual embeddings $\{\mathcal{I}(x)|x\in\mathcal{N}\}$ generated by the image encoder of CLIP for background proposals as in Sec.~\ref{sec:BCP}. 
Then,
we subsequently obtain the embeddings of the cluster centers, denoted as $\{\tilde{\textit{\textbf{w}}}_c|c\in\mathcal{C}_O^{\prime}\}$. These cluster centers function as the embedding centers for the estimated background categories within $\mathcal{C}_O^{\prime}$, allowing for online generation of pseudo labels from background proposals in each training batch.
Throughout the training process, motivated by VL-PLM proposed in \cite{zhao2022exploiting}, background proposals of each training batch undergo being filtered based on RPN scores using a threshold of $\theta$, with an additional step to filter out proposals overlapping with ground-truth boxes. Subsequently, CLIP is employed to generate pseudo labels, preventing the detector from being biased towards estimated background categories.

Then, for a proposal $x\in\mathcal{N}$, its probability score of being classified as $c\in\mathcal{C}_O^{\prime}$ is calculated as follows, 
\begin{equation}
\tilde{p}(c|x; \mathcal{C}_O^{\prime}) = \frac{\exp( \cos(\mathcal{I}(x), \tilde{\textit{\textbf{w}}}_c)/ \tau)}{\sum_{c^{\prime}\in \mathcal{C}_O^{\prime}} \exp(\cos(\mathcal{I}(x). \tilde{\textit{\textbf{w}}}_{c^{\prime}})/ \tau)}.
\end{equation}
Once such background proposal has probability scores corresponding to all categories in $\mathcal{C}_O^{\prime}$, we choose the predicted class label with the highest score as its pseudo label,
\begin{equation}
\textit{\textbf{y}}^o(x)={\underset{c}{\text{argmax}}}\{\tilde{p}(c|x; \mathcal{C}_O^{\prime}) |c\in\mathcal{C}_O^{\prime}\}.
\end{equation}

To eliminate unconfident pseudo labels generated above, we will filter proposals based on probability scores of pseudo labels using a threshold $\theta$. Referring to VL-PLM~\cite{zhao2022exploiting}, we also apply per-class NMS and use RoI Head to refine their box predictions, generating final pseudo labels. 
Afterward, those final pseudo labels are used to assign class labels of $\mathcal{C}_O^{\prime}$ for all background proposals $\mathcal{N}^{B}\subseteq\mathcal{N}$ in each training batch. Let $\mathcal{N}^{B}_p$ denote the set of positive background proposals assigned class labels from $\mathcal{C}_O^{\prime}$, while the remaining proposals are all collected as $\mathcal{N}^{B}_n$.

Therefore, the loss component proposed by this \textbf{BOD} module to optimize the model is formulated as follows,
\begin{equation}
\label{eq:bod}
\begin{split}
\mathcal{L}_{bod}& =  \frac{1}{|\mathcal{N}^{B}_p|} \sum_{x \in \mathcal{N}^{B}_p} - \log p(c=\textit{\textbf{y}}^o(x)|x; \mathcal{C}_{b,o}^{bg})  \\
&+ \lambda_{bg} \frac{1}{|\mathcal{N}^{B}_n|}  \sum_{x \in \mathcal{N}^{B}_n}-\log \sum_{c\in{\mathcal{C}_a}\cup\{\overline{\textit{\textbf{c}}}_{bg}\}} p(c|x; \mathcal{C}_{b,o}^{bg}),
\end{split}
\end{equation}
where $\lambda_{bg}$ is a loss weight with a small value used for background proposals. 
This loss component emphasizes aligning visual embeddings of background proposals with their contextual embeddings related to estimated background categories from $\mathcal{C}_O$. It is especially critical for majority classes within $\mathcal{C}_O$, which are more prone to knowledge loss. Additionally, it empowers \textbf{BOD} to leverage insights from implicit objects within background proposals.

\noindent
\textbf{Training Objective. }
 The final training objective for model optimization is a combination of the loss components formulated by Eq. (\ref{eq:cross_entropy_foreground}), Eq. (\ref{eq:bcp}), and Eq. (\ref{eq:bod}), respectively:
\begin{equation}
\mathcal{L}_{final} = \mathcal{L}_{cls} 
+\mathcal{L}_{bcp}^{\prime} + \mathcal{L}_{bod}.
\end{equation}
It's worth noting that, the text encoder and image encoder of CLIP should be frozen during model training.

\subsubsection{Inference Probability Rectification}


Once performed \textbf{BCP} and \textbf{BOD}, the detector capability of recognizing previously unseen classes has been significantly enhanced.
However, a new challenge arises during inference. The background underlying categories ${\mathcal{C}_O}$, estimated from background proposals during training, might share semantics similarities with the novel classes $\mathcal{C}_u$ of the detector aiming to classify during inference, vividly symbolized by $\mathcal{C}_O \cap \mathcal{C}_u \neq \emptyset$. In such cases, there exist conceptual overlaps of contextual embeddings between those two types of categories during inference. This overlap could hinder the accurate computation of probability scores for novel classes, leading to detection ambiguity during inference. 
To overcome this, an Inference Probability Rectification (\textbf{IPR}) is presented to enable the detector to precisely predict the probabilities of novel classes during inference.

To be specific, let $\mathcal{C}_{b,u,o}^{bg}=\mathcal{C}_b\cup\mathcal{C}_u\cup\mathcal{C}_O\cup\{\overline{\textit{\textbf{c}}}_{bg}\}$. Then, with reference to Eq. (\ref{prob}),
the probability of a proposal $x\in\mathcal{D}_I$ being classified as class $c\in\mathcal{C}_b\cup\mathcal{C}_u$ during inference can be computed as follows,
\begin{equation}
\label{prob_ipr}
p(c|x; \mathcal{C}_{b,u,o}^{bg}) = \frac{\exp( \cos(\textit{\textbf{w}}(x), \textit{\textbf{t}}_{c})/ \tau)}{\mathbf{\Sigma}^{b,u}+\mathbf{\Sigma}^{o}+\mathbf{\Sigma}^{bg}},
\end{equation}
where

\begin{alignat}{2}
&\mathbf{\Sigma}^{b,u} &&= \sum_{c^{\prime}\in \mathcal{C}_b\cup\mathcal{C}_u} \exp(\cos(\textbf{\textit{w}}(x), \textbf{\textit{t}}_{c^{\prime}})/ \tau), 
\nonumber \\
&\mathbf{\Sigma}^{o} &&= \sum_{c^{\prime}\in \mathcal{C}_O} \exp(\cos(\textbf{\textit{w}}(x), \textbf{\textit{t}}_{c^{\prime}})/ \tau), 
\nonumber \\
&\mathbf{\Sigma}^{bg} &&= \sum_{c^{\prime}\in \{\overline{\textit{\textbf{c}}}_{bg}\}} \exp(\cos(\textbf{\textit{w}}(x), \textbf{\textit{t}}_{c^{\prime}})/ \tau). \label{eq:simga_bg}
\end{alignat}


Assuming that a proposal $x$ carries its true class label $c\in\mathcal{C}_u$ simultaneously sharing conceptual overlap with categories in $\mathcal{C}_O$, 
then the probability computation of for this category, namely $p(c|x; \mathcal{C}_{b,u,o}^{bg})$, tends to underestimate its value. This underestimation occurs due to the presence of $c$ simultaneously contributing in both $\mathbf{\Sigma}^{b,u}$ and $\mathbf{\Sigma}^{o}$ in the denominator of Eq. (\ref{prob_ipr}). In essence, $p(c|x; \mathcal{C}_{b,u,o}^{bg})$ becomes smaller than its true probability value, denoted as ${P}(c|x)$, which is a theoretical but unknown value, being discussed later. 
To address this issue, our goal is to resolve the conceptual overlaps between novel categories in $\mathcal{C}_u$ and estimated background underlying categories within $\mathcal{C}_O$. This aims to reduce the influence of $c$ contributing $\mathbf{\Sigma}^{o}$, leading to a re-estimation of $\mathbf{\Sigma}^{o}$ and thus resulting in $\tilde{\mathbf{\Sigma}}^{o}$.


Here, we simplify the function $s(\cdot, \cdot)=\exp(\cos(\cdot, \cdot)/ \tau)$ in Eqs. (\ref{prob_ipr}) to (\ref{eq:simga_bg}) and term its resulting value as the cosine exponential score.
As per the definitions of the softmax function, individual probabilities for each category are computed by normalizing the cosine exponential scores, dividing each by the total sum of cosine exponential scores corresponding to all categories. We hypothesize that the cosine exponential score proportionally reflects its true probability. Hence, $\tilde{\mathbf{\Sigma}}^{o}$ can be represented as:
\begin{equation}
    \tilde{\mathbf{\Sigma}}^{o} = \sum_{c^{\prime}\in {\mathcal{C}}_O} s(\textbf{\textit{w}}(x), \textbf{\textit{t}}_{c^{\prime}}) \cdot ( 1 - \frac{\sum_{c^{\prime\prime}\in {\mathcal{C}}_u}P(c^{\prime}, c^{\prime\prime}|x)}{P(c^{\prime}|x)}). 
    \label{eq:simga_o22}
\end{equation}

According to the Multiplication Rule of Joint Probabilities, the joint probability $P(c^{\prime}, c^{\prime\prime}|x)$ for any $c^{\prime}\in {\mathcal{C}}_O$ and $c^{\prime\prime}\in {\mathcal{C}}_u$ can be estimated as follows:
\begin{equation}
 P(c^{\prime}, c^{\prime\prime}|x) = P(c^{\prime}|x)  P(c^{\prime\prime}|x, c^{\prime}).
 \label{eq:joint}
\end{equation}

With Eq. (\ref{eq:joint}), Eq. (\ref{eq:simga_o22}) can be reformulated as follows:
\begin{equation}
    \tilde{\mathbf{\Sigma}}^{o} = \sum_{c^{\prime}\in {\mathcal{C}}_O} s(\textbf{\textit{w}}(x), \textbf{\textit{t}}_{c^{\prime}}) \cdot ( 1 - \sum_{c^{\prime\prime}\in {\mathcal{C}}_u}P(c^{\prime\prime}|x, c^{\prime})). 
    \label{eq:simga_o23}
\end{equation}

However, estimating $P(c^{\prime\prime}|x, c^{\prime})$ poses challenges. Hence, we assume $ P(c^{\prime\prime}|x, c^{\prime})$ to be sample-agnostic, resulting in $P(c^{\prime\prime}|c^{\prime})$. Given that the embedding space serves as a metric space for probability calculation, $P(c^{\prime\prime}|c^{\prime})$ can be estimated by leveraging the cosine similarity between the contextual embeddings of these two categories as follows,
\begin{equation}
\label{eq:embeddings}
P(c^{\prime\prime}|x, c^{\prime}) = P(c^{\prime\prime}|c^{\prime}) = \frac{ s(\textbf{\textit{t}}_{c^{\prime}}, \textbf{\textit{t}}_{c^{\prime\prime}})}{\sum_{c \in \mathcal{C}_{b,u,o}^{bg}\backslash\{c^{\prime}\}} s(\textbf{\textit{t}}_{c^{\prime}}, \textbf{\textit{t}}_{c})}.
\end{equation}

Considering Eq. (\ref{eq:simga_o23}) and Eq. (\ref{eq:embeddings}) into Eq. (\ref{prob_ipr}),  the probability of a proposal $x\in\mathcal{D}_I$ being classified as class $c\in\mathcal{C}_b\cup\mathcal{C}_u$ during inference can be reformulated as follows,
\begin{equation}
\label{prob_ipr2}
p(c|x; \mathcal{C}_{b,u,o}^{bg}) = \frac{\exp( \cos(\textit{\textbf{w}}(x), \textit{\textbf{t}}_{c})/ \tau)}{\mathbf{\Sigma}^{b,u}+\tilde{\mathbf{\Sigma}}^{o}+\mathbf{\Sigma}^{bg}}.
\end{equation}

\section{Experiments}

\subsection{Experimental Setups}

\noindent
\textbf{Datasets.}
To assess the effectiveness of the proposed LBP framework in handling the OVD task, we conducted experiments on two established object detection benchmark datasets: MS-COCO \cite{lin2014microsoft} and LVIS \cite{gupta2019lvis}. These experiments were conducted in traditional open vocabulary settings \cite{gu2021open, wang2023object, wu2023aligning}, referred to as OV-COCO and OV-LVIS, respectively.
As outlined in prior works \cite{wang2023object, wu2023aligning}, we divided 48 categories as base classes and 17 categories as novel classes in the OV-COCO task. The primary metric used to evaluate the detection performance is the mean average precision at IoU with a threshold of 0.50 (denoted as $\textrm{AP}_{50}$). Specifically, we represent $\textrm{AP}_{50}$ for base and novel categories as $\textrm{AP}^n_{50}$ and $\textrm{AP}^b_{50}$, respectively.
For OV-LVIS, following  \cite{wang2023object, wu2023aligning}, we classified 337 rare categories as novel classes, while considering the remaining common and frequent categories as base classes (resulting in a total of 866 categories). Here, the detector's performance is evaluated using the mean average precision averaged across IoUs from 0.50 to 0.95 (denoted as $\textrm{AP}$). We report $\textrm{AP}$ values for rare categories ($\textrm{AP}_r$), common categories ($\textrm{AP}_c$), frequent categories ($\textrm{AP}_f$), and all classes.
Furthermore, we include experimental results for instance segmentation on LVIS. The metrics $\textrm{AP}^n_{50}$ and $\textrm{AP}_r$ are used as the primary measures to assess the detector's performance on OV-COCO and OV-LVIS, respectively.


\noindent
\textbf{Baselines.}
We compare LBP with the following state-of-the-art (SOTA) algorithms
to handle the OVD task on OV-COCO: Detic~\cite{zhou2022detecting}, Object-centric-OVD~\cite{bangalath2022bridging}, OV-DETR~\cite{zang2022open}, RegionCLIP~\cite{zhong2022regionclip}, ViLD~\cite{gu2021open},
OADP~\cite{wang2023object} and BARON~\cite{wu2023aligning}. Besides, we perform our comparison on  OV-LVIS with previous methods, including ViLD~\cite{gu2021open}, DetPro~\cite{du2022learning} and BARON~\cite{wu2023aligning}.

\subsection{Comparisons with State-of-the-Arts}
Results on OV-COCO and OV-LVIS are reported in Table \ref{tab:OV-COCO} and Table \ref{tab:OV-LVIS}, respectively. 
As shown, our LBP method outperforms previous state-of-the-art (SOTA) algorithms for the OVD task in all cases, validating its effectiveness.
\begin{table}[t]
\centering
\footnotesize
\resizebox{\linewidth}{!}{
\begin{tabular}{l|c|c|ccc}
\toprule
   Method   & Benchmark &  Detector & $\textrm{AP}^n_{50}$ & $\textrm{AP}^b_{50}$ & $\textrm{AP}_{50}$ \\ 
   \midrule
Detic~\cite{zhou2022detecting} &   WS-OVD&   CenterNet2\cite{zhou2021probabilistic}   & ‘   27.8 &  47.1 &  45.0       \\ 
Object-centric-OVD~\cite{bangalath2022bridging}  &   WS-OVD  & Faster R-CNN &  36.6  & 54.0 &49.4 \\   \hline 
VL-PLM    & G-OVD  &   Faster R-CNN  &   32.3 & 54.0 & 48.3        \\   
OV-DETR~\cite{zang2022open}  & G-OVD   &  DeformableDETR\cite{zhu2020deformable} &    29.4 & 61.0 & 52.7 \\ 
\midrule
RegionCLIP \cite{zhong2022regionclip}     &    C-OVD  &  CLIP  & 26.8 & 54.8 & 47.5   \\   
\midrule
ViLD \cite{gu2021open} &   V-OVD       &  Faster R-CNN  &   27.6 & 59.5 &51.3\\     
OADP \cite{wang2023object} &   V-OVD   &   Faster R-CNN     & 30.0  &  53.3 & 47.2 \\ 
BARON \cite{wu2023aligning} &    V-OVD   &   Faster R-CNN   &  34.0  &  60.4  & 53.5      \\ 
BARON$\dagger$  &   V-OVD    &   Faster R-CNN   &  35.8   & 58.2   & 52.3      \\
LBP (ours)  &  V-OVD  &    Faster R-CNN  &  35.9 &  \textbf{60.8}   & \textbf{54.3}  \\ 
LBP$\dagger$ (ours)  &  V-OVD  &    Faster R-CNN  &  \textbf{37.8} &  58.7   &  53.2    \\ 
\bottomrule
\end{tabular}
}
\caption{
Comparison results of LBP and existing SOTA methods on OV-COCO. $\dagger$ indicates model optimization using a batch size of $16$, used for mitigating model overfitting towards base classes. 
}
\label{tab:OV-COCO}
\vspace{-0.1cm}
\end{table}

\begin{table}[t]
\centering
\footnotesize
\resizebox{\linewidth}{!}{
\begin{tabular}{l|cccc|cccc}
\toprule
Method  & \multicolumn{4}{c|}{Object detection} & \multicolumn{4}{c}{Instance segmentation} \\  
      & $\textrm{AP}_r$    & $\textrm{AP}_c$     & $\textrm{AP}_f$     & $\textrm{AP}$     & $\textrm{AP}_r$      & $\textrm{AP}_c$     & $\textrm{AP}_f$     & $\textrm{AP}$       \\ 
\midrule
ViLD \cite{gu2021open}  & 16.7 & 26.5 & \textbf{34.2} & 27.8 & 16.6 & 24.6 & \textbf{30.3} & 25.5     \\         
DetPro \cite{du2022learning}  & 20.8  &  27.8  &  32.4  &  28.4  &  19.8  &  25.6  &  28.9  &  25.9         \\       
BARON \cite{wu2023aligning}  &   20.1  &  28.4 &  32.2   & 28.4  &   19.2   &   26.8   &  29.4  &   26.5 \\ 
$\textrm{BARON} ^\ddagger$&   23.2 & 29.3 & 32.5 & 29.5 & 22.6 & 27.6 & 29.8 & 27.6\\ 
LBP (ours)  & 22.2  &  28.8  &  32.4    &  29.1    &    22.1  &  27.0  &  29.7   &  27.2  \\ 
$\textrm{LBP}^\ddagger$ (ours) & \textbf{24.1} & \textbf{29.5}  &  32.8   &   \textbf{29.9}  &    \textbf{23.7}   &  \textbf{27.7}   &  30.1    &  \textbf{28.0}  \\ 
\bottomrule
\end{tabular}
}
\caption{Comparison results of LBP and existing SOTA methods on OV-LVIS. 
 $\ddagger$ indicates the model trained using learnable prompt templates, proposed by DetPro~\cite{du2022learning}.
}
\label{tab:OV-LVIS}
\end{table}

\noindent
\textbf{Results on OV-COCO.} 
Similar to OADP~\cite{wang2023object}, we categorize the comparison baselines into four OVD benchmark settings: Vanilla OVD (V-OVD), Caption-based OVD (C-OVD), Generalized OVD (G-OVD), and Weakly Supervised OVD (WS-OVD). The emphasis of our proposed approach lies in V-OVD, but we also present performance results across other settings. 
Table \ref{tab:OV-COCO} demonstrates that in this dataset, our LBP approach excels across various benchmark settings, particularly in V-OVD, surpassing existing methods by a notable margin. For instance, compared to BARON$\dagger$, our method demonstrates a 2.0\% improvement in $\textrm{AP}^n_{50}$ and 0.5\% in $\textrm{AP}^b_{50}$ in the V-OVD setting, showcasing its efficacy in detecting both base and novel classes during inference. Moreover, it outperforms previous approaches in other settings, affirming its generality and superiority.


\noindent
\textbf{Results on OV-LVIS.}
Compared to OV-COCO, OV-LVIS poses greater challenges
 due to increased categories and fewer implicit instances for novel classes.
The outcomes in Table~\ref{tab:OV-LVIS} showcase the exceptional performance of our approach. For the task of object detection, our method enhances BARON's performance by 2.1\% on $\textrm{AP}_{r}$ without learnable prompts and by 0.9\% with them. Additionally, our approach consistently boosts the performance of BARON by 2.9\% on $\textrm{AP}_{r}$ in the task of instance segmentation. These comparisons underline the applicability of the proposed method in more complex tasks including object detection and instance segmentation with a larger number of categories, thereby corroborating its effectiveness.

\begin{table}[t]
\centering
\footnotesize
\begin{tabular}{ccc|cccc}
\toprule
 BCP  & BOD & IPR  &  $\textrm{AP}^n_{50}$ & $\textrm{AP}^b_{50}$ & $\textrm{AP}_{50}$  \\ 
\midrule
 -  & -& -   &  35.8  & 58.2 & 52.3 \\
\checkmark   & -& -  & 34.3 & 59.0 & 52.5 \\ 
-   & \checkmark & -  & 35.2 & 58.5 & 52.4 \\ 
\checkmark  & \checkmark& -  & 35.6 & 58.7 & 52.7 \\ 
\checkmark  & - &  \checkmark  & 36.8 & 59.0 & 53.2 \\ 
- & \checkmark  &  \checkmark  & 36.9 & 58.7 & 53.0 \\ 
\checkmark  &  \checkmark  & \checkmark  &  37.8 &  58.7   &  53.2     \\
 \bottomrule
\end{tabular}
\caption{
Ablation study results in individual proposed modules of LBP on OV-COCO.
}
\label{tab:ablation_components}
\end{table}

\subsection{Ablation Analysis}

\noindent
\textbf{Impact of individual proposed modules.}
We conducted an ablation study to assess the effectiveness of each individual module within our method, namely Background Category Prompts (\textbf{BCP}), Background Object Discovery (\textbf{BOD}), and Inference Probability Rectification (\textbf{IPR}). By incrementally integrating these modules into the full model, we sought to understand their individual impact. Table~\ref{tab:ablation_components} illustrates the outcomes. We observed that employing only \textbf{BCP} notably improves the baseline's performance w.r.t. base classes but results in a decline in performance for novel classes. This observation highlights the inherent conceptual overlap generated between the estimated background underlying classes and novel classes when exclusively relying on \textbf{BCP}. However, combining \textbf{BCP} with \textbf{IPR} demonstrates a 2.5\% performance improvement of novel classes, showcasing the effect of \textbf{IPR}. 
Building upon the above variant model, the addition of \textbf{BOD} further enhances the model, resulting in a 1.0\% mAP50 performance improvement on novel classes. This validates its increased role in identifying objects within the estimated underlying categories from background proposals.

\begin{table}[t]
\centering
\footnotesize
\begin{tabular}{l|ccc}
\toprule
Method   &   $\textrm{AP}^n_{50}$ & $\textrm{AP}^b_{50}$ & $\textrm{AP}_{50}$   \\ 
\midrule
Single embedding & 35.8  & 58.2 & 52.3 \\
Soft background loss & 36.4 & 58.4 & 52.6 \\ 
$\mathcal{L}_{bcp}$ & 37.3 & 58.8 & 53.1  \\ 
$\mathcal{L}_{rlx}$ & 36.9 & 58.4 & 52.8\\ 
 LBP (ours) &  37.8 &  58.7   &  53.2  \\ 
 \bottomrule
\end{tabular}
\caption{Additional ablation study results of BCP on OV-COCO.}

\label{tab:ablation_bg}
\end{table}

\noindent
\textbf{Further analysis of $\textbf{BCP}$.}
To further validate $\textbf{BCP}$'s efficacy, we explore various conventional designs for background interpretation in Table~\ref{tab:ablation_bg}. We refer ``Single embedding'' as the design with only a learnable ``background'' embedding $\textbf{\textit{t}}_{bg}$, akin to prior schemes~\cite{gu2021open,wu2023aligning}, while ``Soft background loss'' pertains to the soft background loss outlined in DetPro~\cite{du2022learning}. Subsequently, ``$\mathcal{L}_{bcp}$'' and ``$\mathcal{L}_{rlx}$'' indicate our substitution of $\mathcal{L}_{bcp}^{\prime}$ in 
Eq. (\ref{eq:bcp}) with only $\mathcal{L}_{bcp}$ and $\mathcal{L}_{rlx}$, respectively.
The results considerably highlight the superiority of our proposed design for background interpretation. Specifically, our design significantly outperforms ``Soft background loss'' proposed by DetPro~\cite{du2022learning}, showcasing its effectiveness in preserving class relations. Moreover, the comparison using only $\mathcal{L}_{bcp}$ and $\mathcal{L}_{rlx}$ emphasizes the individual importance of each loss component proposed by $\textbf{BCP}$ used for background interpretation.

\begin{figure}[t]
  \centering

  \begin{subfigure}{0.23\textwidth}
    \centering
    \includegraphics[width=\textwidth, trim=10 15 10 15]{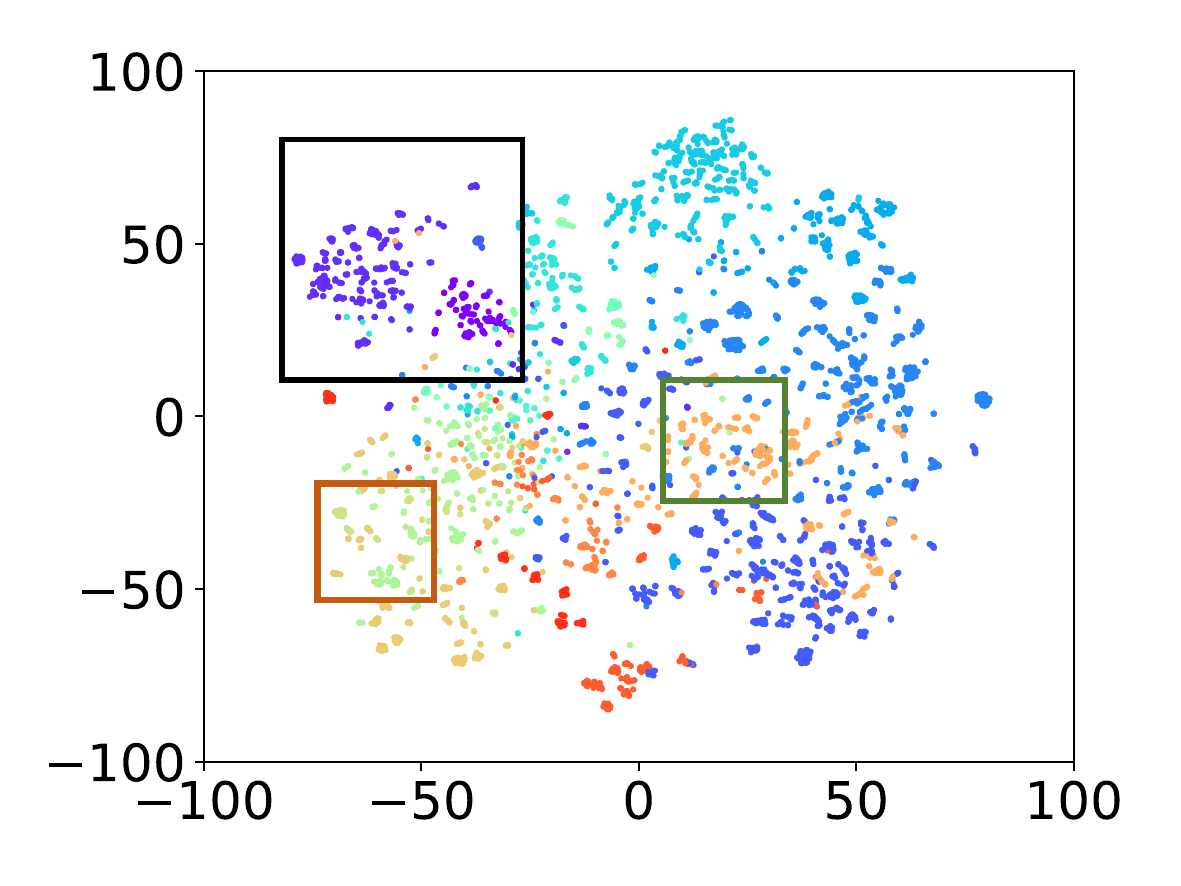}
    \caption{}
    \label{fig:t-SNE-a}
  \end{subfigure}
  \hfill
  \begin{subfigure}{0.23\textwidth}
    \centering
    \includegraphics[width=\textwidth, trim=10 15 10 15]{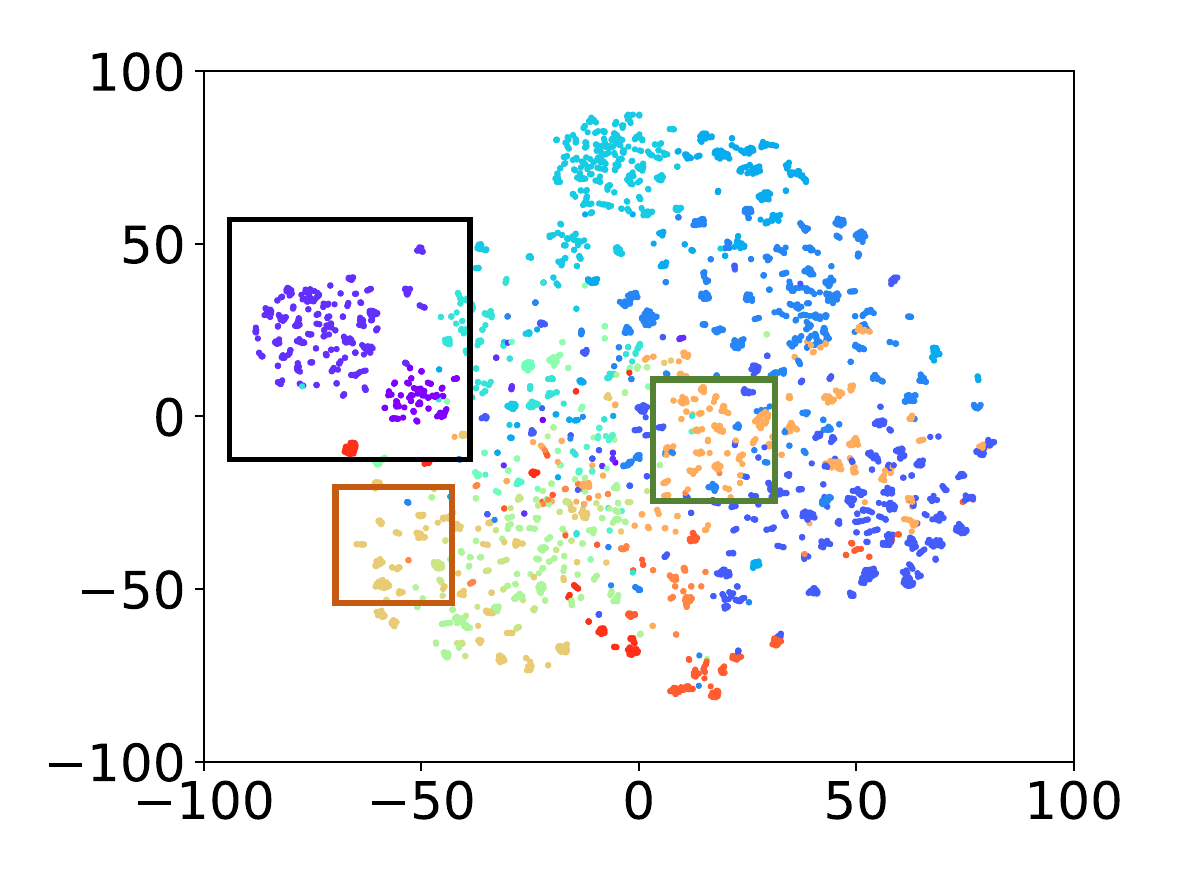}
    \caption{}
    \label{fig:t-SNE-b}
  \end{subfigure}


\vspace{-0.25cm}
  \caption{
        Visualizations of feature distributions for novel category proposals. Note that proposals are selected for those that exhibit significant IoU overlap with the ground truth boxes of novel categories. Different colors denote distinct categories.
    (a) and (b) showcase the feature distributions generated by BARON~\cite{wu2023aligning} and our LBP, respectively. Compared to BARON, our LBP algorithm leads to more compact distributions for the same novel category representations.
}
  \label{fig:2x2subfigures}
\vspace{-0.25cm}
\end{figure}

Furthermore, we use t-SNE to visualize the feature distribution of novel category proposals, further highlighting the efficacy of our designed schemes in background interpretation. 
 As displayed in Figure~\ref{fig:2x2subfigures}(a) and Figure~\ref{fig:2x2subfigures} (b), compared our approach LBP with BARON~\cite{wu2023aligning}, 
the findings illustrate that the proposed LBP approach enables the detector to learn more discriminative features for the proposals associated with novel categories.


\section{Conclusions}

    
    
    
In this paper, we have introduced LBP, a novel framework addressing the challenges of open-vocabulary object detection. 
In this approach, learning background prompts is proposed to harness explored implicit background knowledge. This can enhance the capacity of the detecter to recognize both base and novel categories during inference. 
To achieve this, we have devised three essential modules: Background Category-specific Prompt, Background Object Discovery, and Inference Probability Rectification. These modules collectively empower the detector to discover, represent, and leverage implicit object knowledge explored from background proposals.
Our proposed approach has been rigorously evaluated through extensive experiments and thorough ablation studies, confirming its superior performance.

\section*{Acknowledgments}
This work was supported in part by the National Natural Science Foundation of China (NO.~62322608), in part by the CAAI-MindSpore Open Fund, developed on OpenI Community, in part by the Shenzhen Science and Technology Program (NO.~JCYJ20220530141211024), in part by the Open Project Program of State Key Laboratory of Virtual Reality Technology and Systems, Beihang University (No.VRLAB2023A01). 

{
    \small
    \bibliographystyle{ieeenat_fullname}
    \bibliography{main}
}

\clearpage
\appendix

We present additional implemental details, comparison, and analysis results of the proposed LBP framework in this supplementary material.

\noindent
\section{Implementations.}
We follow the implementation of BARON \cite{wu2023aligning} to conduct experiments. Similar to \cite{radford2021learning,wu2023aligning}, we employ Faster R-CNN \cite{ren2015faster} coupled with ResNet50-FPN \cite{lin2017feature} as the base detector, initializing its backbone network with weights from SOCO \cite{wei2021aligning}. We employ a $2\times$ training schedule (180,000 iterations), with batch size set to 32 (16 for detection and 16 for distillation). We choose SGD \cite{ruder2016overview} as the optimizer, configured with a momentum of 0.90 and a weight decay of $2.50\times10^{-5}$. Additionally, following BARON \cite{wu2023aligning}, we utilize ViT-B/32 CLIP as the PVLM model, with the fixed context prompts from ViLD \cite{gu2021open}. 
 For other hyper-parameters, we maintain consistency across all experiments, such as $n_a=10$, $\theta=0.95$ (consistent with VL-PLM \cite{zhao2022exploiting}), $\tau = 0.02$, $\gamma = 0.02$, and $\lambda_{bg} = 0.05$.

\section{Transfer to Other Datasets}
Similar to \cite{du2022learning,wu2023aligning}, we also report the inference performance of the proposed LBP approach and other previous state-of-the-art (SOTA) OVD methods when transferring a detector trained on the LVIS dataset to three other datasets: Pascal VOC 2007 test set~\cite{everingham2010pascal}, COCOvalidation set~\cite{lin2014microsoft}, and Objects365 v2 validation set~\cite{shao2019objects365}.  As shown in Table~\ref{tab:TRANSFER}, our LBP method demonstrates superior inference performance across these three datasets compared to existing state-of-the-art methods, showcasing the generalized applicability of our LBP approach across various scenarios.
\setcounter{table}{5}
\begin{table*}[t]
\centering
\footnotesize
\begin{tabular}{l|ll|llllll|llllll}
\toprule
      & \multicolumn{2}{c|}{Pascal VOC} & \multicolumn{6}{c|}{MS-COCO} & \multicolumn{6}{c}{Objects365} \\
      &      $\textrm{AP}_{50}$          &      $\textrm{AP}_{75}$       & $\textrm{AP}$   & $\textrm{AP}_{50}$          &      $\textrm{AP}_{75}$ & $\textrm{AP}_{s}$ & $\textrm{AP}_{m}$  & $\textrm{AP}_{l} $ &  $\textrm{AP}$   & $\textrm{AP}_{50}$          &      $\textrm{AP}_{75}$ & $\textrm{AP}_{s}$ & $\textrm{AP}_{m}$  & $\textrm{AP}_{l} $ \\ \midrule
Supervised    &  78.5 & 49.0 & 46.5 & 67.6 & 50.9 & 27.1 & 67.6 & 77.7 & 25.6 & 38.6 & 28.0 & 16.0 & 28.1 &36.7   \\  \midrule
ViLD$\S$ \cite{gu2021open}  &  73.9 & 57.9 & 34.1 & 52.3 & 36.5 & 21.6 & 38.9 & 46.1 & 11.5 & 17.8 & 12.3 & 4.2 & 11.1 & 17.8    \\    
$\textrm{DetPro}$$\S$~\cite{du2022learning}&  74.6 & 57.9 & 34.9 & 53.8 & 37.4 & 22.5 & 39.6 & 46.3 & 12.1 & 18.8 & 12.9 & 4.5 & 11.5 & 18.6    \\ 
$\textrm{BARON}^\ddagger$~\cite{wu2023aligning}&    76.0 & 58.2 & 36.2 & 55.7 & 39.1 & 24.8 & 40.2 & 47.3 & 13.6 & 21.0 & 14.5 & 5.0 & 13.1 & 20.7  \\ 
$\textrm{LBP}^\ddagger$ (ours) &   \textbf{76.1}  &  \textbf{58.4}    &   \textbf{36.8}   &   \textbf{56.5}  &  \textbf{39.8}   &  \textbf{25.6}  & \textbf{40.6} & \textbf{48.1} &  \textbf{14.3} & \textbf{21.8}  &  \textbf{15.1}    &  \textbf{5.5}    & \textbf{13.7}    &  \textbf{21.6} \\

\bottomrule
\end{tabular}
\caption{Comparison results of LBP and existing SOTA methods on Pascal VOC test set, COCO validation set and Object365 validation set with the model being trained on OV-LVIS. 
Specifically, $\S$ indicates those results reported from  DetPro~\cite{du2022learning}, while $\ddagger$ indicates the model trained using learnable prompt templates, proposed by DetPro~\cite{du2022learning}.
}
\label{tab:TRANSFER}
\end{table*}


\section{Additional Ablation Analysis}
\setcounter{figure}{3}
\begin{figure*}[h]
\centering

\includegraphics[width=0.95\linewidth, trim=5 25 25 25, clip]{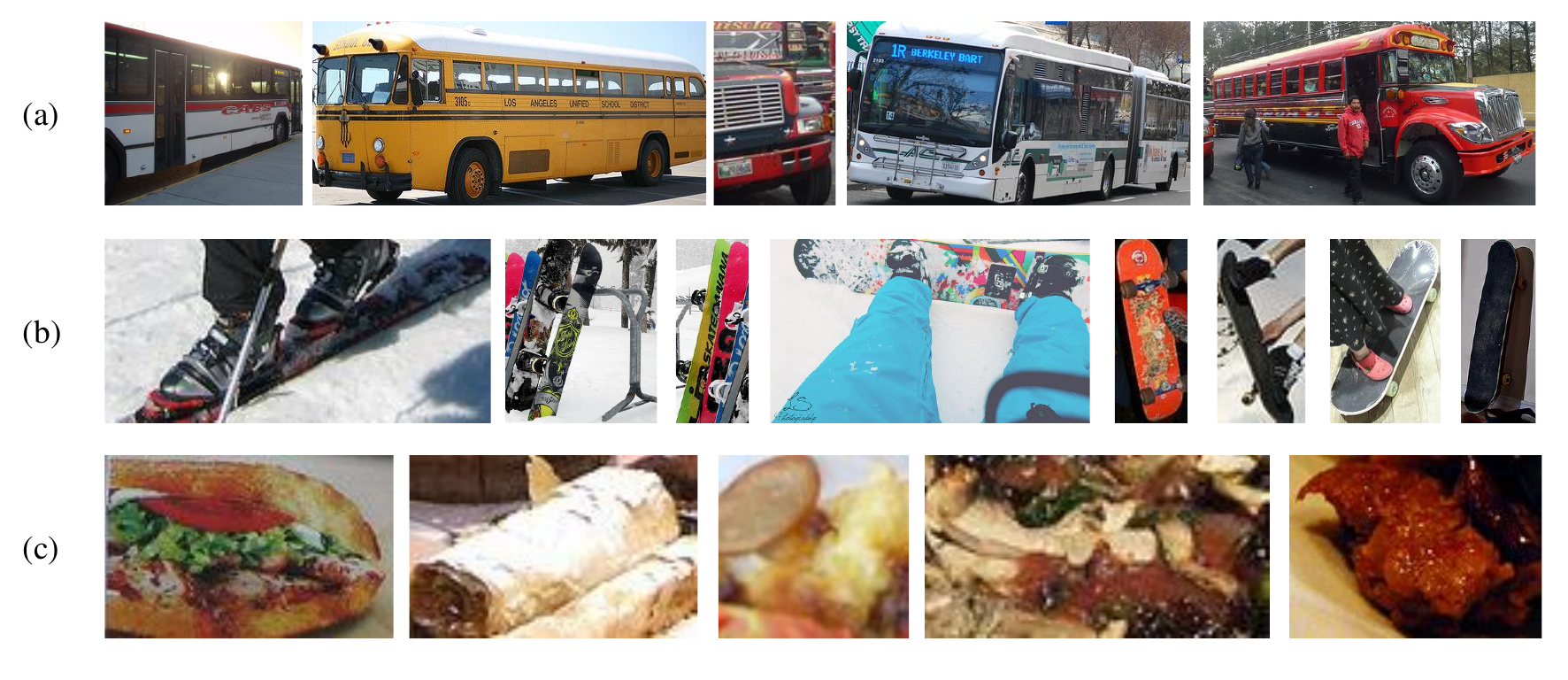}

\caption{Visualizations of representative proposals for three background underlying categories estimated during training. 
They illustrate the potential conceptual overlaps between those background underlying categories and novel categories during inference on OV-COCO.
}
\label{fig:CO}
\end{figure*}
\textbf{Further analysis of BOD.}
To further illustrate the advantages of the proposed \textbf{BOD} module, we attempt to discover representative results to visualize conceptual overlaps between background underlying categories estimated during training and the novel categories during inference, demonstrated in Figure~\ref{fig:CO}. To be more specific, we choose the representative proposals with high predicted probability scores for three background underlying categories.


Figure~\ref{fig:CO}(a) shows one of the estimated background categories, which exhibits significant overlap with ``bus'' proposals belonging to the novel categories. This observation strongly demonstrates the effect of our proposed \textbf{BOD} in discovering meaningful latent categories from a multitude of background proposals during model training.

Conversely,  the other estimated category portrayed in Figure~\ref{fig:CO}(b) demonstrates substantial connections with objects associated with two distinct novel categories: ``skateboard'' and ``snowboard''. This suggests that while \textbf{BOD} might not precisely differentiate all novel categories, it effectively leverages knowledge from visually similar-looking categories, thereby significantly enhancing the model's representations of objects w.r.t. those categories.

In essence, both Figure~\ref{fig:CO}(a) and Figure~\ref{fig:CO}(b) distinctly illustrate conceptual overlaps in contextual embeddings between the background underlying categories estimated during training and the novel categories accessed during inference.

Moreover, Figure~\ref{fig:CO}(c) illustrates that the represented background underlying category encompasses diverse types of food, devoid of significant semantic overlap with the inference novel categories, and nevertheless, it can still detect objects pertinent to that category.
 This discovery underscores the substantial capacities of the proposal model in discovering and leveraging the knowledge of implicit objects from background proposals, markedly bolstering feature discrimination output by our model, and consequently, significantly enhancing detector performance.

\begin{figure}[h]
\centering

\includegraphics[width=0.95\linewidth, trim=25 25 25 25, clip]{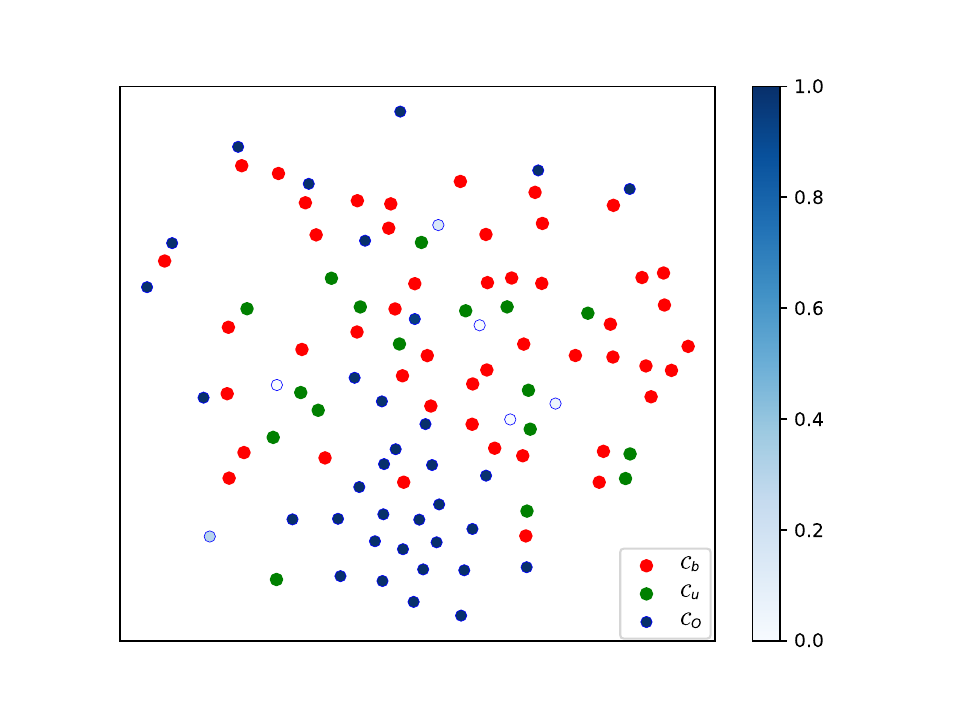}

\caption{
Visualization of the distributions of contextual embeddings of base categories ${\mathcal{C}_b}$, novel categories ${\mathcal{C}_u}$, and background underlying categories ${\mathcal{C}_O}$ in OV-COCO task. 
Here, we harness the magnitude of the shrinking factor, i.e., $( 1 - \sum_{c^{\prime\prime}\in {\mathcal{C}}_u}P(c^{\prime\prime}|x, c^{\prime}))$ in Eq. (\textcolor{red}{18}), to showcase the semantic similarity or conceptual overlap between estimated background categories and inference novel categories. The color bar represents the relationships between $( 1 - \sum_{c^{\prime\prime}\in {\mathcal{C}}_u}P(c^{\prime\prime}|x, c^{\prime}))$ for each $c^{\prime} \in \mathcal{C}_O$ and the shades of blue, where darker shades indicate lower degrees of conceptual overlaps and vice versa.
}
\label{fig:embeddings}
\end{figure}

\noindent
\textbf{More analysis of IPR.}
To further validate the necessity of \textbf{IPR}, we visualize the distributions of the contextual embeddings, encoded by text encoder of CLIP, from base categories ${\mathcal{C}_b}$, novel categories ${\mathcal{C}_u}$, and background underlying categories ${\mathcal{C}_O}$ in the OV-COCO task, depicted in Figure~\ref{fig:embeddings}. As illustrated, during inference, several embeddings from ${\mathcal{C}_O}$ closely resemble those of some novel categories from ${\mathcal{C}_u}$, indicating a probable semantic similarity or conceptual overlap between the two categories. As displayed in Eq. (\textcolor{red}{16}), for each $c^{\prime} \in \mathcal{C}_O$, the proposed \textbf{IPR} module adjusts the cosine exponential score $s(\textbf{\textit{w}}(x), \textbf{\textit{t}}_{c^{\prime}})$ by multiplying it with a shrinking factor, namely $ 1 - \sum_{c^{\prime\prime}\in {\mathcal{C}}_u}P(c^{\prime\prime}|x, c^{\prime})$, to alleviate this issue. Notably, the contextual embeddings of $c^{\prime} \in \mathcal{C}_O$ closer to those of novel categories from ${\mathcal{C}_u}$ exhibit a smaller shrinking factor, demonstrating the effectiveness of our \textbf{IPR} module.

Additionally, compared to conventional designs \cite{gu2021open,wu2023aligning} on background interpretation, Figure~\ref{fig:embeddings} showcases more representation space diversities, conducted by the contextual embeddings of the estimated background underlying categories, further emphasizing the superiority of our LBP approach.


\section{Choices of Hyper-parameters}
To be consistent with VL-PLM~\cite{zhao2022exploiting}, we set $\theta=0.95$ and $\tau = 0.02$ in all experimental cases. In this section, we analyze the choices of other hyper-parameters used in the proposed approach, including $n_a$, $\gamma$ and $\lambda_{bg}$.
\begin{table}[h]
\centering
\footnotesize
\begin{tabular}{c|ccc}
\toprule
$n_a$  &   $\textrm{AP}^n_{50}$ & $\textrm{AP}^b_{50}$ & $\textrm{AP}_{50}$   \\ 
\midrule
$0$ &  37.5 & 58.5  & 53.0  \\ 
$10$ & 37.8 & 58.7 & 53.2\\ 
$20$ & 37.7  & 58.7    & 53.2  \\ 
 \bottomrule
\end{tabular}
\caption{
Ablation study results of our LBP approach under different $n_a$ values  on OV-COCO
}
\label{tab:ablation_na}
\end{table}

\begin{table}[h]
\centering
\footnotesize
\begin{tabular}{c|ccc}
\toprule
$\gamma$  &   $\textrm{AP}^n_{50}$ & $\textrm{AP}^b_{50}$ & $\textrm{AP}_{50}$   \\ 
\midrule
$0.01$ & 37.6 & 58.8 & 53.2  \\ 
$0.02$ & 37.8 & 58.7 & 53.2\\ 
$0.05$ & 37.5  & 58.7  & 53.1  \\ 
 \bottomrule
\end{tabular}
\caption{
Ablation study results of our LBP approach under different $\gamma$ values 
 on OV-COCO.
}
\label{tab:ablation_gamma}
\end{table}

\begin{table}[h]
\centering
\footnotesize
\begin{tabular}{c|ccc}
\toprule
$\lambda_{bg}$  &   $\textrm{AP}^n_{50}$ & $\textrm{AP}^b_{50}$ & $\textrm{AP}_{50}$   \\ 
\midrule
$0.01$ & 37.4 & 58.9 & 53.3  \\ 
$0.05$ & 37.8 & 58.7 & 53.2\\ 
$0.10$ &  37.8 &  58.6  & 53.1 \\ 
 \bottomrule
\end{tabular}
\caption{
Ablation study results of our LBP approach under different $\lambda_{bg}$ values on OV-COCO.
}
\label{tab:ablation_lambda}
\end{table}

\noindent\textbf{Choice of $n_a$.} 
To determine $n_a$, we compared model performance under different $n_a$ values on OV-COCO, as outlined in Table~\ref{tab:ablation_na}. Here, $n_a=10$ represents our default setting, acting as the baseline among its variants. When $n_a=0$, indicating no expansion of estimated background categories, the model's performance decreased by 0.3\% in $\textrm{AP}^n_{50}$ and 0.5\% in $\textrm{AP}^b_{50}$ compared to the baseline. This vividly demonstrates the effectiveness of our proposed strategy to expand estimated background categories. On the other hand, increasing $n_a$ to 20 results in a 0.1\% decrease in performance in $\textrm{AP}^n_{50}$ compared to the baseline, suggesting that a larger $n_a$ may do harm to further improve the performance of the model.

\noindent\textbf{Choice of $\gamma$.}
Table~\ref{tab:ablation_gamma} illustrates the detector performance of our LBP approach under different $\gamma$ values. The results indicate that our method is not overly sensitive to the choice of $\gamma$. Setting $\gamma$ to $0.01$ or $0.05$, as opposed to the default 0.02, only leads to a slight decrease in the detection performance, w.r.t. novel categories.

\noindent\textbf{Choice of $\lambda_{bg}$}.
To better select $\lambda_{bg}$, we present its performance under different settings in Table~\ref{tab:ablation_lambda}. As illustrated, when $\lambda_{bg}=0.05$, the model achieves the best performance compared to that other values of $\lambda_{bg}$. Hence, we set it as the default and consider it the baseline among different variations. Specifically, reducing $\lambda_{bg}$ to 0.01 resulted in a 0.4\% decrease in model performance in $\textrm{AP}^n_{50}$ compared to the baseline. This indirectly indicates that setting a larger $\lambda_{bg}$ can prevent the detector from overfitting to background underlying categories $\mathcal{C}_O$.

\end{document}